\documentclass{article}
\PassOptionsToPackage{numbers, sort&compress, square}{natbib}

\usepackage[final]{neurips_2019}

\usepackage[utf8]{inputenc} 
\usepackage[T1]{fontenc}    
\usepackage{hyperref}       
\usepackage{url}            
\usepackage{booktabs}       
\usepackage{amsfonts}       
\usepackage{nicefrac}       
\usepackage{microtype}      

\usepackage{graphicx}
\usepackage{subcaption}
\usepackage{multirow}
\usepackage{setspace}
\usepackage{amsmath,amssymb}
\usepackage{mathtools}
\usepackage{multicol}
\usepackage{float}
\usepackage{verbatim}
\usepackage{algorithm}
\usepackage{algorithmic}
\usepackage{color}
\usepackage[font=small]{caption}
\usepackage{wrapfig}

\usepackage{amsmath,amsfonts,bm}

\def\1{\bm{1}}

\DeclareMathAlphabet{\mathsfit}{\encodingdefault}{\sfdefault}{m}{sl}
\SetMathAlphabet{\mathsfit}{bold}{\encodingdefault}{\sfdefault}{bx}{n}

\DeclareMathOperator*{\argmax}{arg\,max}

\makeatletter
\usepackage{xspace}
\def\@onedot{\ifx\@let@token.\else.\null\fi\xspace}
\DeclareRobustCommand\onedot{\futurelet\@let@token\@onedot}

\newcommand{\figref}[1]{Fig\onedot~\ref{#1}} 
\newcommand{\equref}[1]{Eq\onedot~\eqref{#1}}
\newcommand{\secref}[1]{Sec\onedot~\ref{#1}}
\def\eg{\emph{e.g}\onedot} 
\def\ie{\emph{i.e}\onedot}

\def\wrt{w.r.t\onedot}

\title{Graph Structured Prediction Energy Networks}

\author{%
  Colin Graber \hspace{40pt} Alexander Schwing \\
  {\small \texttt{cgraber2@illinois.edu} \hspace{20pt} \texttt{aschwing@illinois.edu}}\\
  Department of Computer Science\\
  University of Illinois at Urbana-Champaign\\
  Champaign, IL \\
  \\
}

\begin{document}

\maketitle

\vspace{-0.8cm}
\begin{abstract}

For joint inference over multiple variables, a variety of structured prediction
techniques have been developed to  model  correlations among variables and
thereby 
improve predictions. However, many classical approaches  suffer from one of two
primary drawbacks: they either lack the ability to model high-order correlations
among variables while maintaining computationally tractable inference, or they
do not allow to explicitly model  known correlations. To address this
shortcoming, we introduce `Graph Structured Prediction Energy Networks,' for which we develop inference techniques
that allow to both model explicit local  and  implicit
higher-order correlations while maintaining tractability of inference. We apply
the proposed method to  tasks from the natural
language processing and computer vision domain and demonstrate its general utility.

\end{abstract}

\vspace{-0.3cm}
\section{Introduction}
\vspace{-0.2cm}
Many machine learning tasks involve joint prediction of a set of variables. For instance, semantic image segmentation infers the class label for every pixel in an image.  
To address joint prediction, it is common to use  deep nets which model probability
distributions independently over the variables (\eg, the pixels). 
The downside:  correlations  between different variables aren't modeled explicitly. 

A number of techniques, such as Structured SVMs~\citep{Tsochantaridis2005}, Max-Margin Markov Nets~\citep{Taskar2003} and Deep Structured Models~\citep{ChenSchwingICML2015,SchwingARXIV2015}, directly model relations between output variables. However, modeling the correlations between a large number of variables is  computationally expensive and therefore generally impractical. 
As an attempt to address some of the fallbacks of classical high-order structured prediction techniques, Structured Prediction Energy Networks (SPENs) were introduced \citep{BelangerICML2016,belanger2017end}. SPENs assign a score to an entire prediction, which allows them to harness global structure. Additionally, because these models do not represent structure explicitly, complex relations between variables can be learned while maintaining tractability of inference. However, SPENs have their own set of downsides: \citet{BelangerICML2016} mention, and we can confirm, that it is easy to overfit SPENs to the training data. Additionally, the inference techniques developed for SPENs do not enforce structural constraints among output variables,  hence they cannot support structured scores and discrete  losses. An attempt to combine locally structured scores  with joint prediction  was introduced very recently by \citet{graber2018nlstruct}. However,~\citet{graber2018nlstruct} require the score function to take a specific, restricted form, and inference is formulated as a difficult-to-solve saddle-point optimization problem.

To address these concerns, we develop a new model which we refer to as `Graph Structured Prediction Energy Network' (GSPEN). Specifically, GSPENs combine the capabilities of classical structured prediction models and SPENs and have the ability to explicitly model local structure when  known or assumed, while  providing the ability to learn an unknown or more global structure implicitly. Additionally, the proposed GSPEN formulation generalizes the approach by~\citet{graber2018nlstruct}. Concretely, inference in GSPENs is a maximization of a generally non-concave function \wrt structural constraints, for which we develop two inference algorithms.

We show the utility of GSPENs by comparing  to related techniques on several tasks: optical character recognition, image tagging, multilabel classification, and named entity recognition. In general, we show that GSPENs are able to outperform other models. Our implementation is available at \url{https://github.com/cgraber/GSPEN}.

\vspace{-0.2cm}
\section{Background}
\label{sec:bg}
\vspace{-0.2cm}
Let $x \in \mathcal{X}$ represent the input provided to a model, such as a sentence or an image. In this work, we consider tasks where the outputs take the form $y = (y_1, \dots, y_K) \in \mathcal{Y} \coloneqq \prod_{k=1}^K \mathcal{Y}_k$, \ie, they are vectors  where the $k$-th variable's domain is the discrete and finite set $\mathcal{Y}_k = \{1, \ldots, |\mathcal{Y}_k|\}$. In general, the number of variables $K$ which are part of the configuration $y$ can depend on the observation $x$. However, for readability only, we  assume all $y \in \mathcal{Y}$ contain $K$ entries, \ie, we drop the dependence of the output space $\mathcal{Y}$ on input $x$. 

All models we consider consist of a function $F(x, y;w)$, which assigns a score to a given configuration $y$ conditioned on input $x$ and is parameterized by weights $w$. Provided an input $x$, the inference problem requires finding the configuration $\hat y$ that maximizes this score, \ie, $\hat{y} \coloneqq \argmax_{y \in \mathcal{Y}} F(x, y; w)$. 

To find the parameters $w$ of the function $F(x,y;w)$, it is common to use a Structured Support Vector Machine (SSVM) (a.k.a.\ Max-Margin Markov Network) objective \citep{Tsochantaridis2005, Taskar2003}: given a multiset $\left\{\left(x^i, y^i\right)_{i=1}^N\right\}$ of data points $\left(x^i,y^i\right)$ comprised of an input $x^i$ and the corresponding ground-truth configuration $y^i$, a SSVM attempts to find weights $w$ which maximize the margin between the scores assigned to the ground-truth configuration $y^i$ and the inference prediction:
\begin{equation}
	\min_w \sum_{\left(x^i, y^i\right)} \max_{\hat{y} \in \mathcal{Y}}\left\{ F\left(x^i, \hat{y}; w\right) + L\left(\hat{y}, y^i \right) \right\} - F\left(x^i, y^i; w\right).
	\label{eq:SSVM}
\end{equation}
Hereby, $L\left(\hat{y}, y^i\right)$ is a task-specific and often discrete loss, such as the Hamming loss, which  steers the model towards learning a margin between  correct and incorrect outputs. Due to addition of the task-specific loss $L(\hat y, y^i)$ to the model score $F\left(x^i,\hat y;w\right)$, we often refer to the maximization task within \equref{eq:SSVM} as loss-augmented inference. The  procedure  to solve  loss-augmented inference  depends on the considered model, which we discuss next.

\textbf{Unstructured Models.}
Unstructured models, such as  feed-forward deep nets, assign a score to each label of variable $y_k$ which is part of the configuration $y$, irrespective of the label choice of other variables. Hence, the final score function $F$ is the sum of $K$ individual scores $f_k(x, y_k; w)$, one for each variable:
\begin{equation}
	F(x, y; w) \coloneqq \sum_{k=1}^K f_k(x, y_k; w).
\end{equation}
Because the scores for each output variable do not depend on the scores assigned to other output variables, the inference assignment is determined efficiently by 
independently finding the maximum score for each variable $y_k$. The same is true for loss-augmented inference, assuming that the loss decomposes into a sum of independent terms as well.
 
\textbf{Classical Structured Models.}
\label{sec:struct}
Classical structured models incorporate dependencies between variables by considering functions that take more than one output space variable $y_k$ as input, \ie, each  function depends on a subset $r \subseteq \{1, \dots, K\}$ of the output variables.  We refer to the subset of variables via $y_r = (y_k)_{k\in r}$ and use $f_r$ to denote the corresponding function.   The  overall score for a configuration $y$ is a sum of these functions, \ie, 
\begin{equation}
	F(x, y;w) \coloneqq \sum_{r \in \mathcal{R}} f_r(x, y_r; w).
\end{equation}
	Hereby, $\mathcal{R}$ is a set containing all of the variable subsets which are required to compute $F$. The variable subset relations between functions $f_r$, \ie, the structure, is often visualized   using factor graphs or, generally, Hasse diagrams.
	
This formulation allows to explicitly model relations between variables, but it
comes at the price of more complex inference which is
NP-hard~\citep{Shimony1994} in general. A number of approximations to this problem have
been developed and utilized successfully (see  \secref{sec:related} for more
details), but the complexity of these methods scales with the size of the
largest region $r$. For this reason, these models  commonly consider only unary
and pairwise regions, \ie, regions with one or two variables.
	
Inference, \ie, maximization of the score, is equivalent to the integer linear program 
\begin{equation}
	\max_{p \in \mathcal{M}} \sum_{r \in \mathcal{R}} \sum_{y_r \in \mathcal{Y}_r} p_r(y_r)f_r(x, y_r; w),
\end{equation}
where each $p_r$ represents a marginal probability vector for region $r$ and $\mathcal{M}$ represents the set of $p_r$ whose marginal distributions are globally consistent, which is often called the marginal polytope. 
Adding an entropy term over the probabilities to the inference objective  transforms the problem from maximum a-posteriori (MAP)  to marginal inference, and pushes the predictions to be more uniform \citep{Wainwright2008,NiculaeMBC18}. When combined with the learning procedure specified above, this entropy provides learning with the additional interpretation of maximum likelihood estimation \citep{Wainwright2008}. The training objective then also fits into the framework of Fenchel-Young Losses \citep{blondel2018learning}.

For computational reasons, it is common to relax the marginal polytope $\mathcal{M}$ to the local polytope $\mathcal{M}_L$, which is the set of all probability vectors that marginalize consistently for the factors present in the graph \citep{Wainwright2008}. Since the resulting marginals are no longer globally consistent, \ie, they are no longer guaranteed to arise from a single joint distribution, we write the predictions for each region using $b_r(y_r)$ instead of $p_r(y_r)$ and refer to them using the term ``beliefs.'' Additionally, the entropy term is approximated using fractional entropies~\cite{Heskes2003} such that it only depends on the factors in the graph, in which case it takes the form $H_{\mathcal{R}}(b) \coloneqq \sum_{r\in \mathcal{R}} \sum_{y_r \in \mathcal{Y}_r} - b_r(y_r) \log b_r(y_r)$.

\textbf{Structured Prediction Energy Networks.}
\label{sec:spen}
Structured Prediction Energy Networks (SPENs)~\citep{BelangerICML2016} were motivated by the desire to represent interactions between larger sets of output variables without incurring a high computational cost. The SPEN score function takes the following form: 
\begin{equation}
F(x, p_1, \dots, p_K; w) \coloneqq T\left(\bar{f}(x; w), p_1, \dots, p_K; w \right),
\end{equation}
where $\bar{f}(x; w)$ is a learned feature representation of the input $x$, each $p_k$ is a one-hot vector, and $T$ is a function that takes these two terms and assigns a score. This representation of the labels, \ie, $p_k$, is used to facilitate  gradient-based optimization during inference. More specifically,  inference is formulated via the  program:
\begin{equation}
	\max_{p_{k} \in \Delta_k \forall k} T\left( \bar{f}(x; w), p_1, \dots, p_K; w \right),
\end{equation}
where each $p_{k}$ is constrained to lie in the $|\mathcal{Y}_k|$-dimensional probability simplex $\Delta_k$. This task can be solved using any constrained optimization method. However, for non-concave $T$ the inference solution might only be approximate.

\textbf{NLStruct.}
\label{sec:nlstruct}
SPENs do not support score functions that contain a structured component. 
In response, \citet{graber2018nlstruct} introduced NLStruct, which combines a classical structured score function with a nonlinear transformation applied on top of it to produce a final score. Given a set $\mathcal{R}$ as defined previously, the NLStruct score function takes the following form:
\begin{equation}
	F(x,p_\mathcal{R};w) \coloneqq T\left(f_\mathcal{R}(x; w) \circ p_\mathcal{R}; w \right),
\end{equation}
where $f_\mathcal{R}(x; w) \coloneqq (f_r(x, y_r; w))|_{r \in \mathcal{R}, y_r \in \mathcal{Y}_r}$ is a vectorized form of the score function for a classical structured model, $p_\mathcal{R} \coloneqq (p_r(y_r))|_{\forall r \in \mathcal{R}, \forall y_r \in \mathcal{Y}_r}$ is a vector containing all  marginals, `$\circ$' is the Hadamard product, and $T$ is a scalar-valued function. 

For this model, inference is formulated as a constrained optimization problem, where $\mathcal{Y}_\mathcal{R} \coloneqq \prod_{r \in \mathcal{R}}\mathcal{Y}_r$: 
\begin{equation}
	\max_{y \in \mathbb{R}^{|\mathcal{Y}_\mathcal{R}|}, p_\mathcal{R} \in \mathcal{M}} T(y; w) \text{~~s.t.~~} y = f_\mathcal{R}(x; w) \circ p_\mathcal{R}. 
\end{equation}
Forming the Lagrangian of this program and rearranging leads to the saddle-point inference problem
\begin{equation}
	\min_{\lambda} \max_y \left\{T(y; w) - \lambda^T y \right\} + \max_{p_\mathcal{R} \in \mathcal{M}} \lambda^T \left(f_\mathcal{R}(x; w) \circ p_\mathcal{R}\right).
\end{equation}
Notably,  maximization over $p_\mathcal{R}$ is solved using techniques developed
for classical structured models\footnote{As mentioned,  
solving the maximization over $p_\mathcal{R}$ tractably might require relaxing the marginal polytope $\mathcal{M}$ to the local marginal polytope $\mathcal{M}_L$. For brevity, we will not repeat this fact whenever an inference problem of this form appears throughout the rest of this paper.}, and the saddle-point problem is optimized using the primal-dual algorithm of \citet{chambolle2011first}, which alternates between updating $\lambda$, $y$, and $p_\mathcal{R}$.

\vspace{-0.2cm}
\section{Graph Structured Prediction Energy Nets}
\label{sec:ours}
\vspace{-0.2cm}
Graph Structured Prediction Energy Networks (GSPENs) generalize all
aforementioned models. They combine 
both a classical
structured component as well as a SPEN-like component to score an entire set
of predictions jointly. Additionally, the GSPEN score function  is
more general than that for NLStruct, and includes it as a special case. After
describing the formulation of both the score function and the inference problem
(\secref{sec:model}), we discuss two  approaches to solving inference
(\secref{sec:fw} and \secref{sec:md}) that we found to work well in practice.
Unlike the methods described previously for NLStruct, these approaches do not
require solving a saddle-point optimization problem.

\vspace{-0.2cm}
\subsection{GSPEN Model}
\label{sec:model}
\vspace{-0.2cm}
The GSPEN score function is written as follows:
\begin{equation*}
	F\left(x, p_\mathcal{R}; w\right) \coloneqq T\left(\bar{f}(x; w), p_\mathcal{R}; w \right),
\end{equation*}
where vector $p_{\mathcal{R}} \coloneqq (p_r(y_r))|_{r \in \mathcal{R}, y_r \in \mathcal{Y}_r}$ contains one marginal per region per assignment of values to that region. This formulation allows for the  use of a structured score function while also allowing $T$ to score an entire prediction jointly. Hence, it is a combination of classical structured models and SPENs. For instance, we can construct a GSPEN model by summing a classical structured model and a multilayer perceptron that scores an entire label vector, in which case the score function  takes the form $F(x,p_\mathcal{R};w) \coloneqq \sum_{r \in \mathcal{R}} \sum_{y_r \in \mathcal{Y}_r} p_r(y_r)f_r(x, y_r; w) + \text{MLP}\left(p_\mathcal{R}; w\right)$. Of course, this is one of many possible score functions that are supported by this formulation. Notably, we recover the NLStruct score function if we use $T(\bar{f}(x; w), p_\mathcal{R}; w) = T'(\bar{f}(x; w) \circ p_\mathcal{R}; w)$ and let $\bar{f}(x; w) = f_\mathcal{R}(x; w)$.

Given this model, the inference problem is
\begin{equation}
\label{eq:hspen_inf}
	\max_{p_\mathcal{R} \in \mathcal{M}} T\left(\bar{f}(x; w), p_\mathcal{R}; w\right).
\end{equation}
As for classical structured models, the probabilities  are
constrained to lie in the marginal polytope. In addition 
we also consider  a fractional entropy term over the predictions,
leading to 
\begin{equation}
	\max_{p_\mathcal{R} \in \mathcal{M}}  T\left(\bar{f}(x; w), p_\mathcal{R}; w\right) + H_\mathcal{R}(p_\mathcal{R}).
	\label{eq:hspen_inf_h}
\end{equation}
In the classical setting, adding an entropy term 
relates to Fenchel duality~\citep{blondel2018learning}. However, the GSPEN inference objective does not take the correct form to use this reasoning. 
We instead view this entropy as  a regularizer for the predictions: it pushes
predictions towards a uniform distribution, smoothing the inference
objective, which we empirically observed to improve convergence. The 
results discussed below indicate that adding
entropy   leads to  better-performing models. Also
note that it is possible to add a similar entropy term to the SPEN inference
objective, which is mentioned by~\citet{BelangerICML2016} and~\citet{belanger2017end}. 
 
For inference in GSPEN, SPEN  procedures cannot be used  since they do not maintain the additional constraints imposed by the graphical model, \ie, the marginal polytope ${\cal M}$. We also cannot use the inference procedure developed for NLStruct, as the GSPEN score function does not take the same form. 
 Therefore, in the following, we describe two inference algorithms that optimize
 the program while maintaining structural constraints. 

\vspace{-0.2cm}
\subsection{Frank-Wolfe Inference}
\label{sec:fw}
\vspace{-0.2cm}

\begin{figure}[t]
\vspace{-0.4cm}
\begin{minipage}[t]{0.49\linewidth}
\begin{algorithm}[H]
	\caption{Frank-Wolfe Inference for GSPEN\\}
		\label{alg:fw}
\begin{algorithmic}[1]
	\STATE {\bfseries Input:} Initial set of predictions $p_\mathcal{R}$; Input $x$; \\
	\quad\quad\quad Factor graph $\mathcal{R}$\\
	\FOR{$t = 1 \dots T$}
		
		\STATE $g \Leftarrow \nabla_{p_\mathcal{R}}F(x, p_\mathcal{R};w)$
		\STATE $\widehat{p}_\mathcal{R} \Leftarrow \max\limits_{\widehat{p}_\mathcal{R} \in \mathcal{M}_\mathcal{R}} \sum\limits_{r \in \mathcal{R},y_r \in \mathcal{Y}_r} \hat{p}_r(y_r) g_r(y_r)$ 
		\STATE $p_\mathcal{R} \Leftarrow p_\mathcal{R} + \frac{1}{t}\left(\widehat{p}_\mathcal{R} - p_\mathcal{R}\right)$\vspace{0.015cm}
		
	\ENDFOR
	\STATE {\bfseries Return:} $p_\mathcal{R}$
\end{algorithmic}
\end{algorithm}
\end{minipage}
\begin{minipage}[t]{0.5\linewidth}
\begin{algorithm}[H]
	\caption{Structured Entropic Mirror Descent Inference}
	\label{alg:md}
\begin{algorithmic}[1]
	\STATE {\bfseries Input:} Initial set of predictions $p_\mathcal{R}$; Input $x$; \\
	\quad \quad\quad Factor graph $\mathcal{R}$\\
	\FOR{$t = 1 \dots T$}
		\STATE $g \Leftarrow \nabla_{p_\mathcal{R}}F\left(x, p_\mathcal{R};w\right)$
		\STATE $a \Leftarrow 1 + \ln p_\mathcal{R} + g/\sqrt{t}$
		\STATE \hspace{-0.1cm}$p_\mathcal{R} \hspace{-0.1cm}\Leftarrow\hspace{-0.1cm} \max\limits_{\hat{p}_\mathcal{R} \in \mathcal{M}} \hspace{-0.00cm}\sum\limits_{r \in \mathcal{R},y_r \in \mathcal{Y}_r} \hspace{-0.4cm}\hat{p}_r(y_r) a_r(y_r) \!\!+\!\! H_\mathcal{R}(\hat{p}_\mathcal{R})$
	\ENDFOR
	\STATE {\bfseries Return:} $p_\mathcal{R}$
\end{algorithmic}
\end{algorithm}
\end{minipage}
\vspace{-0.5cm}
\label{fig:no1}
\end{figure}

The Frank-Wolfe algorithm \citep{Frank1956} is suitable because the objectives in Eqs.~(\ref{eq:hspen_inf},~\ref{eq:hspen_inf_h}) are  non-linear while the constraints are linear. 
Specifically, using \citep{Frank1956}, we
compute a linear approximation of the objective at the current iterate, maximize this linear approximation subject to the constraints of the original problem, and take a step towards this maximum.  

In Algorithm~\ref{alg:fw} we detail the steps to optimize \equref{eq:hspen_inf}. In every iteration we first calculate the gradient of the score function $F$ with respect to the marginals/beliefs using the  current prediction as input. We denote this gradient using $g = \nabla_{p_\mathcal{R}} T\left(\bar{f}(x; w), p_\mathcal{R}; w \right)$. The gradient of $T$ depends on the specific function used and is computed via backpropagation. If entropy is part of the  objective, an additional term of $- \ln\left(p_\mathcal{R}\right)-1$ is added to this gradient. 

Next we find the maximizing beliefs which  
is equivalent to  inference  for classical structured prediction: the constraint
space is identical and the objective is a linear function of the
marginals/beliefs. Hence, we solve this inner optimization  using
one of a number of techniques referenced in  \secref{sec:related}.

Convergence guarantees for Frank-Wolfe have been proven when the overall
objective is concave, continuously differentiable, and has bounded curvature~\citep{jaggi2013revisiting}, which is the case when $T$ has these properties with respect to the marginals. This is true even when the inner optimization  is only
solved approximately, which is often  the case due to  standard
approximations used for structured inference. When $T$ is non-concave,
convergence can still be guaranteed, but only to a local
optimum~\citep{lacoste2016convergence}. Note that entropy has unbounded curvature, therefore its inclusion in the objective precludes convergence guarantees. Other variants of the
Frank-Wolfe algorithm exist which  improve convergence in certain
cases~\citep{krishnan2015barrier,lacoste2015global}. We defer a study of these properties to future work.

\vspace{-0.2cm}
\subsection{Structured Entropic Mirror Descent}
\label{sec:md}
\vspace{-0.2cm}
Mirror descent, another constrained optimization algorithm, is analogous to
projected subgradient descent, albeit using a more general  distance beyond
the Euclidean one~\citep{beck2003mirror}. This algorithm has been used in the
past to solve inference for SPENs, where entropy was used as the link function
$\psi$ and by normalizing over each coordinate independently~\citep{BelangerICML2016}. We similarly use entropy in our case. However, the
additional constraints  in  form of the polytope ${\cal M}$ require special care. 

\begin{figure*}[t]
\vspace{-0.2cm}
\centering
\fbox{
\begin{minipage}{0.973\textwidth}
\begin{equation*}
\min_w \sum_{\left(x^{(i)}, p_\mathcal{R}^{(i)}\right)} \left[ \max_{\hat{p}_\mathcal{R} \in \mathcal{M}}\left\{ T\left(\bar{f}(x; w), \hat{p}_\mathcal{R}; w\right) + L\left(\hat{p}_\mathcal{R}, p_\mathcal{R}^{(i)}\right) \right\} - T\left(\bar{f} \left(x^{(i)}; w\right), p_\mathcal{R}^{(i)}; w\right) \right]_+
\label{eq:learn}
\end{equation*}
\end{minipage}
}
\vspace{-0.2cm}
\caption{The GSPEN learning formulation, consisting of a Structured SVM (SSVM) objective with loss-augmented inference. Note that each $p_\mathcal{R}^{(i)}$ are one-hot representations of labels $y_i$. }
\label{fig:learn}
\vspace{-0.6cm}
\end{figure*}

We summarize the structured entropic mirror descent inference  for the proposed
model in Algorithm~\ref{alg:md}.  Each iteration of mirror descent updates the
current prediction $p_\mathcal{R}$ and dual vector $a$ in two  steps: (1) $a$ is
updated based on the current prediction $p_\mathcal{R}$. Using
$\psi(p_\mathcal{R}) = - H_\mathcal{R}(p_\mathcal{R})$ as the link function,
this update step takes the form $a = 1 + \ln p_\mathcal{R} +
\frac{1}{\sqrt{t}}\left(\nabla_{p_\mathcal{R}} T\left(\bar{f}(x;
w),p_\mathcal{R}; w \right) \right)$. As mentioned previously, the gradient of
$T$ can be computed using backpropagation; (2) $p_\mathcal{R}$ is updated by
computing the maximizing argument of the Fenchel conjugate of the link function
$\psi^*$ evaluated at $a$. More specifically, $p_\mathcal{R}$ is updated via
\begin{equation}
	p_\mathcal{R} = \max_{\hat{p}_\mathcal{R} \in \mathcal{M}} \sum_{r \in \mathcal{R}} \sum_{y_r \in \mathcal{Y}_r} \hat{p}_r(y_r) a_r(y_r) + H_\mathcal{R}\left(\hat{p}_\mathcal{R}\right),
\end{equation} 
which is identical to classical structured prediction.

When the inference objective is concave and Lipschitz continuous (\ie, when $T$ has these properties), this
algorithm has also been proven to converge~\citep{beck2003mirror}.
Unfortunately, we are not aware of any convergence results if the inner
optimization problem is solved approximately and if the objective is not
concave. In practice, though, we did not observe any convergence issues during
experimentation.

\vspace{-0.2cm}
\subsection{Learning GSPEN Models}
\label{sec:train}
\vspace{-0.2cm}
 GSPENs   assign a score to an input $x$ and a prediction $p$. 
An SSVM learning objective is applicable,  which maximizes the margin between
the scores assigned to the correct prediction and the inferred result. The full SSVM
learning objective with added loss-augmented inference is summarized in
\figref{fig:learn}. The learning procedure consists of computing the
highest-scoring prediction using one of the inference procedures described in
\secref{sec:fw} and \secref{sec:md} for each example in a mini-batch and then
updating the weights of the model towards making better predictions.

\section{Experiments}
\vspace{-0.2cm}
\begin{wraptable}{r}{0.51\textwidth}
\centering
\setlength{\tabcolsep}{2pt}
{\small
\begin{tabular}{lcccc}
\toprule
& Struct & SPEN & NLStruct & GSPEN\\
\midrule
OCR (size 1000) & 0.40 s &  0.60 s & 68.56 s & 8.41 s \\
Tagging & 18.85 s & 30.49 s & 208.96 s & 171.65 s \\
Bibtex & 0.36 s &  11.75 s & -- & 13.87 s \\
Bookmarks & 6.05 s & 94.44 s & -- & 234.33 s\\
NER & 29.16 s & -- & -- & 99.83 s\\
\bottomrule
\end{tabular}
}
\vspace{-0.2cm}
\caption{ 
Average time to compute  inference objective and complete a weight update for one pass through the training data.
We show all  models trained for this work. 
}
\label{tab:time}
\vspace{-0.3cm}
\end{wraptable}

To demonstrate the utility of our model and to compare  inference and learning settings, we report  results on the tasks of optical character recognition (OCR), image tagging, multilabel classification, and named entity recognition (NER). For each experiment, we use the following baselines: Unary is an unstructured model that does not explicitly model the correlations between output variables in any way. Struct is a classical deep structured model using neural network potentials. We  follow the inference and learning formulation of~\citep{ChenSchwingICML2015}, where inference consists of a message passing algorithm derived using block coordinate descent on a relaxation of the inference problem.  SPEN and NLStruct represent the formulations discussed in  \secref{sec:spen}. Finally, GSPEN represents Graph Structured Prediction Energy Networks, described in  \secref{sec:ours}. For GSPENs, the inner structured inference problems are solved using the same algorithm as for Struct. To compare the run-time of these approaches, Table~\ref{tab:time} gives the average epoch compute time (\ie, time to compute the inference objective and update model weights) during training for our models for each task. In general, GSPEN training was more efficient with respect to time than NLStruct but, expectedly, more expensive than SPEN. Additional experimental details, including hyper-parameter settings, are provided  in Appendix~\ref{sec:exp_details}.

\vspace{-0.2cm}
\subsection{Optical Character Recognition (OCR)}
\vspace{-0.2cm}

\begin{wrapfigure}{r}{0.5\textwidth}
\vspace{-1.2cm}
\centering
\includegraphics[width=0.5\textwidth]{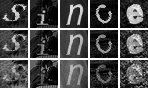}
\vspace{-0.6cm}
\caption{OCR sample data points with different interpolation factors $\alpha$. }
\label{fig:ocr_sample}
\vspace{-0.5cm}
\end{wrapfigure}
For the OCR experiments, we generate data by  selecting a list of 50 common 5-letter English words, such as `close,' `other,' and `world.' To create each data point, we choose a  word from this list and render each letter as a 28x28 pixel image by selecting a random image of the letter from the Chars74k dataset~\citep{de2009character}, randomly shifting, scaling,  rotating, and interpolating  with a random background image patch.  A different pool of backgrounds and letter images was used for the training, validation, and test splits of the data. The  task is to identify the words given 5 ordered images. We create three versions of this dataset using different interpolation factors of $\alpha \in \{0.3, 0.5, 0.7\}$, where each pixel in the final image is computed as $\alpha x_{\text{background}} + (1-\alpha)x_{\text{letter}}$. See  \figref{fig:ocr_sample} for a sample from each dataset. This process was deliberately designed to ensure that information about the structure of the problem (\ie, which words exist in the data) is a strong signal, while the signal provided by each individual letter image can be adjusted. The training, validation, and test set sizes for each dataset are 10,000, 2,000, and 2,000, respectively. During training we vary the training data to be either 200, 1k or 10k.

\begin{figure}[t]
\vspace{-0.3cm}
\centering
{\small
\begin{subfigure}[t]{0.33\textwidth}
\centering
\includegraphics[width=\textwidth]{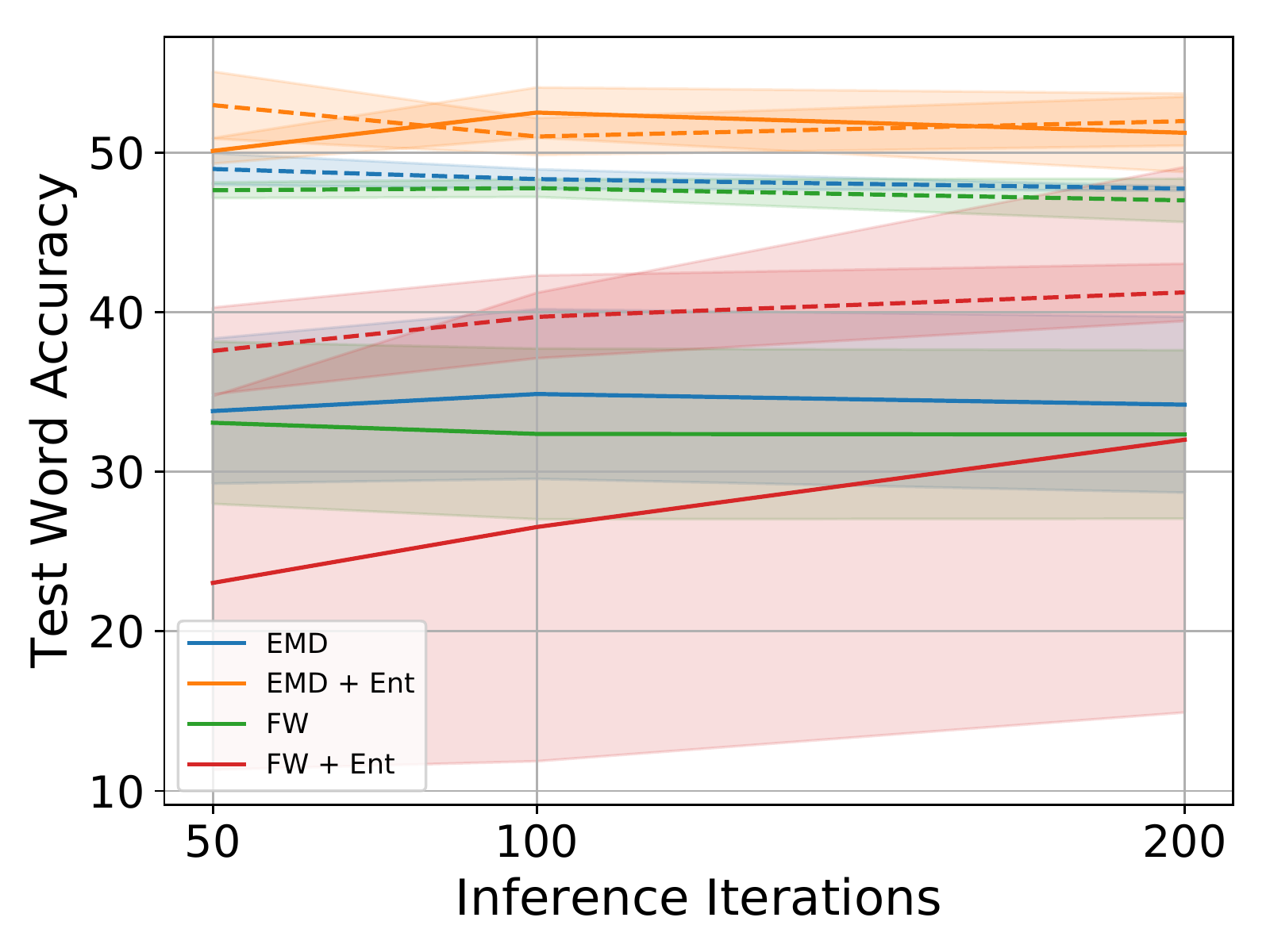}
\vspace{-0.6cm}
\subcaption{}
\label{fig:ocr_inf}
\end{subfigure}~
\begin{subfigure}[t]{0.33\textwidth}
\centering
\includegraphics[width=\textwidth]{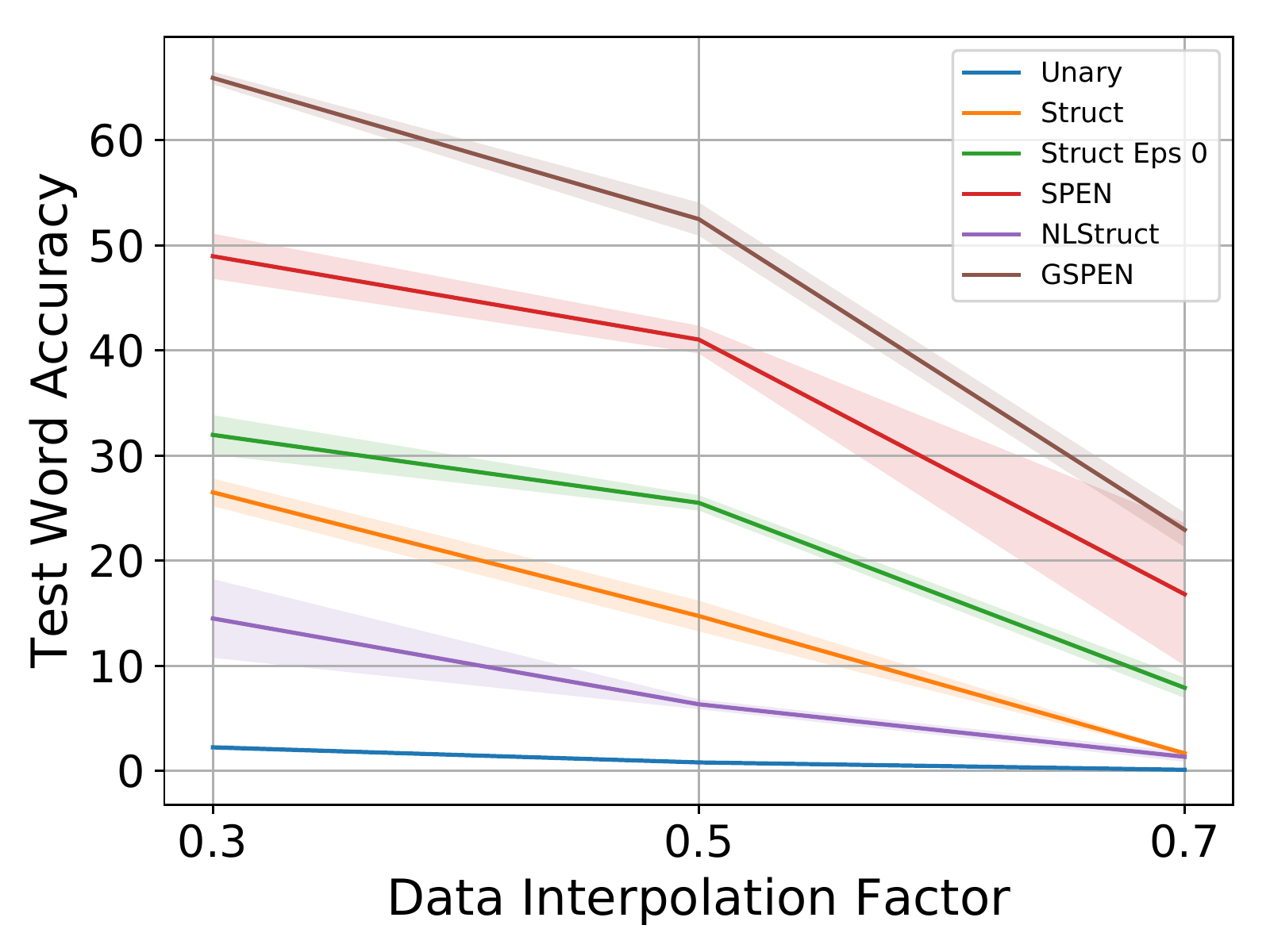}
\vspace{-0.6cm}
\subcaption{}
\label{fig:ocr_interp}
\end{subfigure}~
\begin{subfigure}[t]{0.33\textwidth}
\centering
\includegraphics[width=\textwidth]{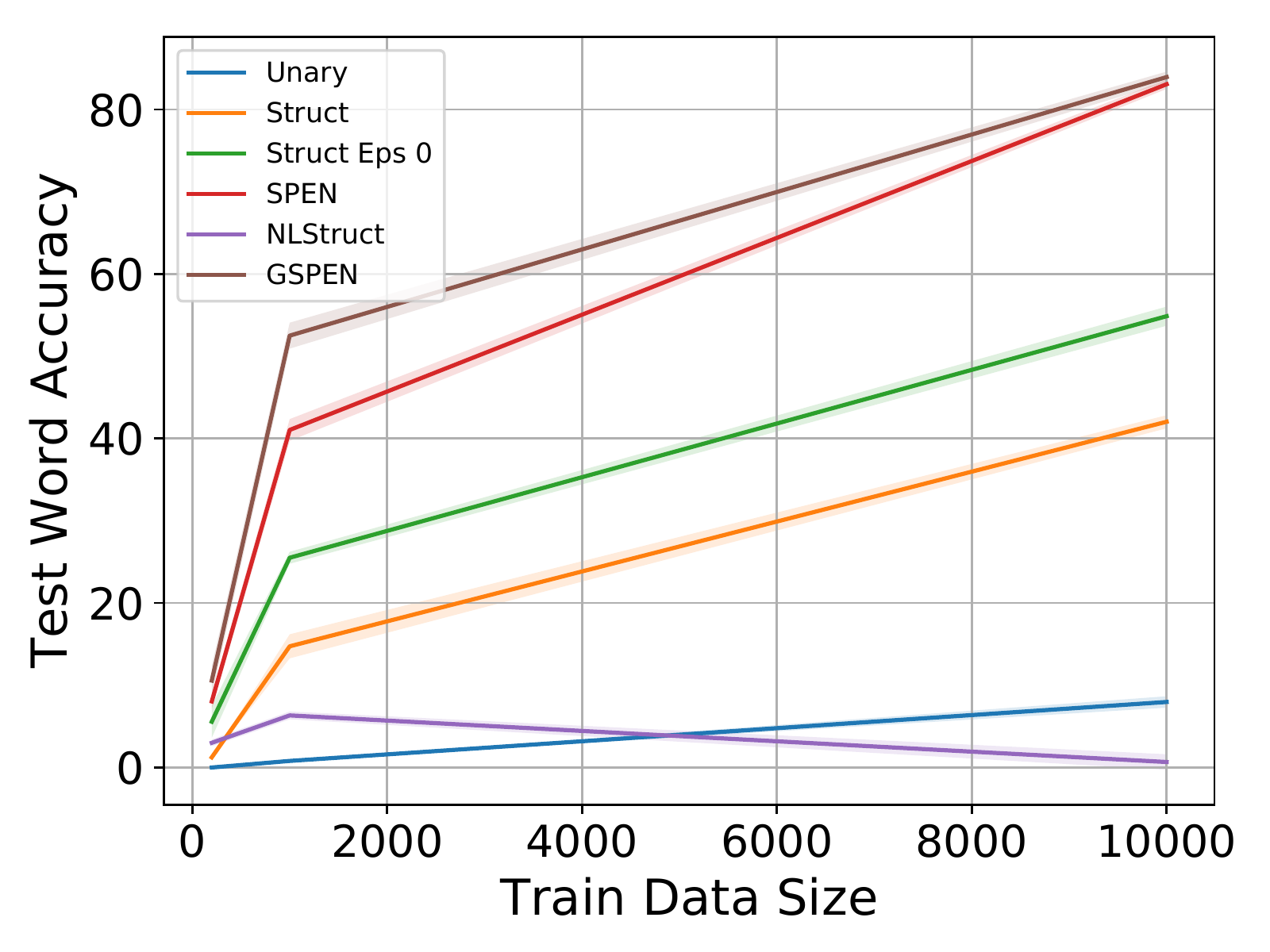}
\vspace{-0.6cm}
\subcaption{}
\label{fig:ocr_data}
\end{subfigure}
}
\vspace{-0.3cm}
\caption{Experimental results on OCR data. The dashed lines in (a) represent models trained from Struct without entropy, while solid lines represent models trained from Struct with entropy. }
\vspace{-0.6cm}
\end{figure}

To study  the inference algorithm, we train four different GSPEN models on the dataset containing 1000 training points and using $\alpha=0.5$. Each model uses either Frank-Wolfe or Mirror Descent and included/excluded the entropy term. To maintain tractability of inference, we fix a maximum iteration count for each model. We additionally investigate the effect of this maximum count on final performance. Additionally, we run this experiment by initializing from two different Struct models,  one being trained using entropy during inference and one being trained without entropy. The results for this set of experiments are shown in \figref{fig:ocr_inf}. Most configurations perform similarly across the number of iterations, indicating these choices are sufficient for convergence. When initializing from the models trained without entropy, we observe that both Frank-Wolfe without entropy and Mirror Descent with entropy performed comparably. When initializing from a model trained with entropy, the use of mirror descent with entropy led to much better results. 

The results for all values of $\alpha$ using a train dataset size of 1000 are presented in \figref{fig:ocr_interp}, and results for all train dataset sizes with $\alpha=0.5$ are presented in \figref{fig:ocr_data}. We observe that, in all cases, GSPEN outperforms all baselines. The degree to which GSPEN outperforms other models depends most on the amount of train data: with a sufficiently large amount of data, SPEN and GSPEN perform comparably. However, when less data is provided, GSPEN performance does not drop as sharply as that of SPEN initially. It is also worth noting that GSPEN outperformed NLStruct by a large margin. The NLStruct model is less stable due to its  saddle-point formulation. Therefore it is much harder to obtain good performance with this model.

\vspace{-0.2cm}
\subsection{Image Tagging}
\vspace{-0.2cm}

Next, we evaluate on the MIRFLICKR25k dataset \citep{huiskes2008mir}, which consists of 25,000 images taken from Flickr. Each image is assigned a subset of  24 possible tags. The train/val/test sets for these experiments consist of 10,000/5,000/10,000 images, respectively.

We compare to NLStruct and SPEN. 
We initialize the structured portion of our GSPEN model using the pre-trained DeepStruct model described by \citet{graber2018nlstruct}, which consists of unary potentials produced from an AlexNet architecture \citep{KrizhevskyNIPS2012} and linear pairwise potentials of the form $f_{i,j}(y_i, y_j, W) = W_{i,j,x_i,x_j}$, \ie,  containing one weight per pair in the graph per assignment of values to that pair. A fully-connected pairwise graph was used. The $T$ function for our GSPEN model consists of a 2-layer MLP with 130 hidden units. It takes as input a concatenation of the unary potentials generated by the AlexNet model and the current prediction. Additionally, we train a SPEN model with the same number of layers and hidden units. We used Frank-Wolfe without entropy for both SPEN and GSPEN inference. 

\begin{wrapfigure}{r}{0.5\textwidth}
\vspace{-0.6cm}
\includegraphics[width=0.5\textwidth]{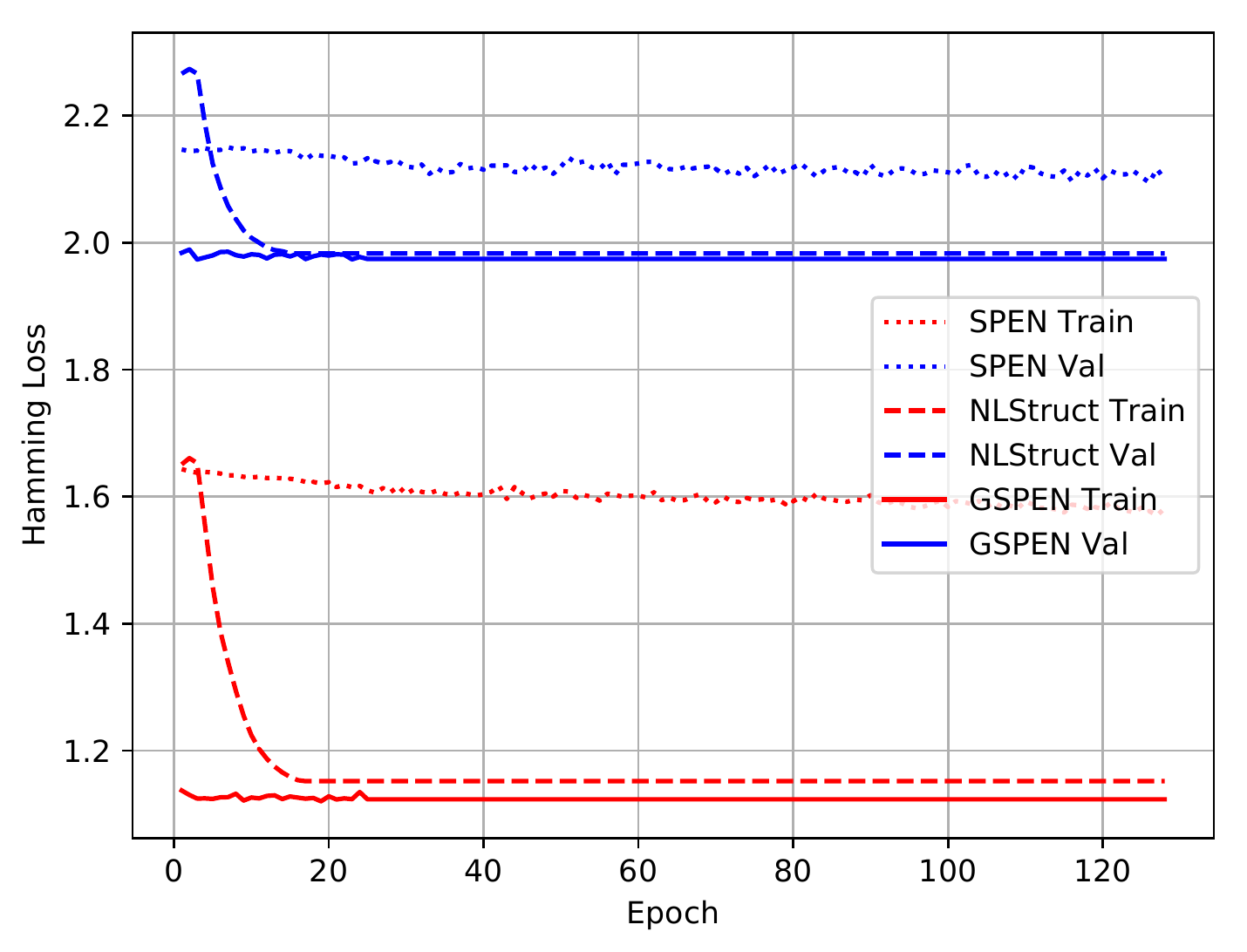}
\vspace{-0.7cm}
\caption{Results for image tagging.}
\label{fig:tagging_results}
\vspace{-0.5cm}
\end{wrapfigure}
The results are shown in \figref{fig:tagging_results}. GSPEN obtains similar test performance to the NLStruct model, and both outperform SPEN. However, the NLStruct model was run for 100 iterations during inference without reaching `convergence' (change of objective smaller than threshold), while the GSPEN model required an average of 69 iterations to converge at training time and 52 iterations to converge at test time. Our approach has the advantage of requiring fewer variables to maintain during inference and requiring fewer iterations of inference  to converge. The final test losses for SPEN, NLStruct and GSPEN are 2.158, 2.037, and 2.029, respectively.

\vspace{-0.2cm}
\subsection{Multilabel Classification}
\vspace{-0.2cm}

We use the Bibtex and Bookmarks multilabel datasets \citep{katakis2008multilabel}. 
They consist of binary-valued input feature vectors, each of which is assigned some subset of 159/208 possible labels for Bibtex/Bookmarks, respectively. We train unary and SPEN models with  architectures identical to \citep{BelangerICML2016} and \citep{GygliICML2017} but add dropout layers. In addition, we further regularize the unary model by flipping each bit of the input vectors with probability 0.01 when sampling mini-batches. For Struct and GSPEN, we generate a graph by first finding the label variable that is active in  most training data label vectors  and  add edges connecting every other  variable to this most active one. Pairwise potentials are generated by passing the input vector through a 2-layer MLP with 1k hidden units. The GSPEN model is trained by starting from the SPEN model, fixing its parameters, and training the pairwise potentials.

The results  are  in Table~\ref{tab:mlc} alongside those taken from  \citep{BelangerICML2016} and \citep{GygliICML2017}. We found  the Unary models to  perform similarly to or better than previous best results. Both SPEN and Struct are able to improve upon these Unary  results. GSPEN outperforms all  configurations, suggesting that the contributions of the SPEN component and the Struct component to the score function are complementary.

\vspace{-0.2cm}
\subsection{NER}
\vspace{-0.2cm}
We also assess suitability for  
 Named Entity Recognition (NER) using the English portion of the CoNLL 2003 shared task \citep{tjong2003introduction}. To demonstrate the applicability of GSPEN for this task, we transformed two separate models, specifically the ones presented by \citet{ma2016end} and \citet{akbik2018contextual}, into GSPENs by taking their respective score functions and adding a component that jointly scores an entire set of predictions. In each case, we first train six instances of the structured model using different random initializations and drop the model that performs the worst on validation data. We then train the GSPEN model, initializing the structured component from these pre-trained models. 

The final average performance is presented in Table~\ref{tab:ner}, and individual trial information can be found in Table~\ref{tab:ner_full} in the appendix. When comparing to the model described by \citet{ma2016end},  GSPEN improves the final test performance in four out of the five trials, and GSPEN has a higher overall average performance across both validation and test data. Compared to \citet{akbik2018contextual}, on average GSPEN's validation score was higher, but it performed slightly worse at test time. Overall, these results demonstrate that it is straightforward to augment a task-specific structured model with an additional prediction scoring function which can lead to improved final task performance.

\begin{table}[t]
\vspace{-0.4cm}
\centering
\begin{tabular}{cc}
\begin{minipage}[t]{0.49\textwidth}
\begin{table}[H]
\caption{Multilabel classification  results for all models. All entries represent macro F1 scores. The top results are taken from the cited publications.}
\label{tab:mlc}
\vspace{-0.12cm}
\setlength{\tabcolsep}{5pt}
\begin{center}
{\small
\begin{tabular}{lcccc}\toprule
&\multicolumn{2}{c}{Bibtex} & \multicolumn{2}{c}{Bookmarks}\\
& Validation & Test & Validation & Test\\
\midrule
SPEN \citep{BelangerICML2016} & -- & 42.2 & -- & 34.4\\
DVN \citep{GygliICML2017}  & -- & 44.7 & -- & 37.1 \\
\midrule
Unary & 43.3 & 44.1  & 38.4 & 37.4 \\
Struct &  45.8 &  46.1 & 39.7 & 38.9 \\
SPEN & 46.6 & 46.5 &  40.2 & 39.2\\
GSPEN & \textbf{47.5}  & \textbf{48.6} & \textbf{41.2} & \textbf{40.7} \\\bottomrule
\end{tabular}
}
\end{center}
\end{table}
\end{minipage}
&
\begin{minipage}[t]{0.46\textwidth}
\begin{table}[H]
\caption{Named Entity Recognition results for all models. All entries represent F1 scores averaged over five trials.}
\label{tab:ner}
\vspace{-0.1cm}
\begin{center}
{\small
\begin{tabular}{rccc}
\toprule
&  Avg. Val. & Avg. Test\\\\
\midrule
Struct \citep{ma2016end} &   94.88 $\pm$ 0.18 & 91.37 $\pm$ 0.04 \\
+ GSPEN & \textbf{94.97} $\pm$ 0.16 & \textbf{91.51} $\pm$ 0.17 \\
\midrule
Struct \citep{akbik2018contextual} &    95.88 $\pm$ 0.10 & \textbf{92.79} $\pm$ 0.08\\
+ GSPEN  & \textbf{95.96} $\pm$ 0.08 & 92.69 $\pm$ 0.17\\
\bottomrule
\end{tabular}
}
\vspace{-0.7cm}
\end{center}
\end{table}
\end{minipage}
\end{tabular}
\vspace{-0.6cm}
\end{table}

\vspace{-0.2cm}
\section{Related Work}
\label{sec:related}
\vspace{-0.2cm}
A variety of techniques have been developed to model  structure among output variables, originating from seminal  works of \citep{Lafferty2001,Taskar2003,Tsochantaridis2005}. These works focus on extending linear classification, both probabilistic and non-probabilistic, to  model the correlation among output variables. Generally speaking, 
scores representing both predictions for individual output variables and for combinations of output variables are used. A plethora of techniques have been developed to solve inference for problems of this form, \eg,~\cite{Schlesinger1976, Werner2007, Boykov1998, Boykov2001, Wainwright2003b, Globerson2006, Welling2004, Sontag2012, Batra2011, Sontag2008, Sontag2007, Wainwright2008, Sontag2009, Murphy1999, Meshi2009, Globerson2007, Wainwright2005b, Wainwright2005, Wainwright2003, Heskes2006, Hazan2010, Hazan2008, Yanover2006, Meltzer2009, Weiss2007, Heskes2002, Heskes2003, Yedidia2001, Ihler2004, Wiegerinck2003, SchwingNIPS2012, SchwingICML2014, SchwingCVPR2011a, Komodakis2010, MeshiNIPS2015, MeshiNIPS2017}. As exact inference for general structures is NP-hard \citep{Shimony1994}, early work focused on   tractable exact inference. However, due to interest in modeling problems with intractable structure, a plethora of approaches have been studied for learning with
approximate inference \citep{Finley2008, Kulesza2008, Pletscher2010, Hazan2010b, Meshi2010, komodakis2011efficient, Schwing2011a, MeshiICML2016}. More recent work has also investigated the role of different types of prediction regularization, with \citet{NiculaeMBC18} replacing the often-used entropy regularization with an L2 norm and \citet{blondel2018learning} casting both   as special cases of a  Fenchel-Young loss framework. 

To model both non-linearity and structure, 
deep learning and structured prediction techniques were combined. Initially,  local, per-variable score functions were learned with deep nets and  correlations among output variables were learned in a separate second stage \citep{AlvarezECCV2012, ChenICLR2015}. Later work simplified this process, learning both local score functions and variable correlations jointly \citep{TompsonNIPS2014, zheng2015conditional, ChenSchwingICML2015, SchwingARXIV2015, LinNIPS2015}. 

Structured Prediction Energy Networks (SPENs), introduced by \citet{BelangerICML2016}, take a different approach to modeling structure. Instead of explicitly specifying a structure a-priori and enumerating scores for every assignment of labels to regions, SPENs learn a function which assigns a score to an input and a label. Inference uses  gradient-based optimization to maximize the score \wrt the label. \citet{belanger2017end} extend this technique by
unrolling inference in a manner inspired by \citet{domke2012generic}. Both  approaches involve iterative inference procedures, which are  slower than feed-forward prediction of deep nets. To improve inference speed, \citet{tu2018learning} learn a neural net  to produce the same output as the gradient-based  methods. Deep Value Networks \citep{GygliICML2017} follow the same approach of \citet{BelangerICML2016} but use a different  objective that encourages the score  to equal the task loss of the prediction. All  these approaches   do not permit to include known structure. The proposed approach enables this. 

Our  approach is most similar to our earlier work \cite{graber2018nlstruct}, which combines explicitly-specified structured potentials with a SPEN-like score function. The score function of our earlier work is a special case of the one presented here. 
In fact, earlier we required a classical structured prediction model as an intermediate layer of the score function, while we  don't make this assumption any longer. Additionally, in our earlier work we had to solve inference via 
a computationally challenging saddle-point objective. Another related approach is described by \citet{vilnis2015bethe}, whose score function is the sum of a classical structured score function and a (potentially non-convex) function of the marginal probability vector $p_\mathcal{R}$. This is also a special case of the score function presented here. Additionally, the inference algorithm they develop is based on regularized dual averaging \citep{xiao2010dual} and takes advantage of the structure of their specific score function, \ie, it is not directly applicable to our setting.

\vspace{-0.cm}
\section{Conclusions}
\vspace{-0.cm}
The developed GSPEN model combines the strengths of several prior approaches to solving structured prediction problems. It allows machine learning practitioners to include inductive bias in the form of known structure into a model while implicitly capturing higher-order correlations among output variables. The model formulation described here is more general than previous attempts to combine explicit local and implicit global structure modeling while not requiring inference to solve a saddle-point problem. 

\subsubsection*{Acknowledgments}
This work is supported in part by NSF under Grant No.\ 1718221 and MRI \#1725729, UIUC, Samsung, 3M, Cisco Systems Inc.\ (Gift Award CG 1377144) and Adobe. We thank NVIDIA for providing GPUs used for this work and Cisco for access to the Arcetri cluster.

{\small
\bibliography{refs,AlexAll}
\bibliographystyle{unsrtnat}
}
\clearpage
\appendix
\section{Appendix}
In this Appendix, we first present additional experimental results for the NER  before providing additional details on experiments, including model architectures and selection of hyper-parameters.

\subsection{Additional Experimental Results}
\begin{table}
\caption{All Trials for NER experiments.}
\label{tab:ner_full}
\begin{center}
\begin{tabular}{ccc|ccc}
\toprule
& \multicolumn{2}{c}{\citet{ma2016end}} & \multicolumn{2}{c}{+ GSPEN }\\
Trial & Val. F1 & Test F1 & Val. F1 & Test F1\\
\midrule
 $1$ & $94.94$ & $91.36$ & $94.85$ & $91.25$\\
$2$ & $94.67$ & $91.37$  & $94.76$ & $91.53$\\
$3$ & $94.74$ & $91.32$  & $95.07$ & $91.60$\\
$4$ & $94.94$ & $91.35$  & $95.08$ & $91.47$\\
$5$ & $95.12$ & $91.44$  & $95.10$ & $91.71$\\
\midrule
& \multicolumn{2}{c}{\citet{akbik2018contextual}} & \multicolumn{2}{c}{+ GSPEN }\\
\midrule
$1$ & $0.9576$ & $0.9284$ & $0.9587$ & $0.9288$\\
$2$ & $0.9601$ & $0.9277$ & $0.9599$ & $0.9246$\\
$3$ & $0.9581$ & $0.9271$ & $0.9588$ & $0.9271$\\
$4$ & $0.9592$ & $0.9291$ & $0.9600$ & $0.9280$\\
$5$ & $0.9590$ & $0.9272$ & $0.9604$ & $0.9260$\\
\bottomrule
\end{tabular}
\end{center}
\end{table}
Table~\ref{tab:ner_full} contains the results for all trials of the NER experiments. When comparing to  \citet{ma2016end}, we outperform their model on both validation and test data in four out of five trials. When comparing to \citet{akbik2018contextual}, we outperform their model on validation data in four out of five trials, but only outperform their model on test data in one trial.

\subsection{Additional Experimental Details}
\label{sec:exp_details}
\paragraph{General Details:}
Unless otherwise specified, all Struct models were trained by using the corresponding pre-trained Unary model, fixing these parameters, and training pairwise potentials. All SPEN models were trained by using the pre-trained Unary model, fixing these parameters, and training the $T$ function. Early stopping based on task performance on validation was used to select the number of epochs for training. For SPEN, GSPEN, and Struct models, loss-augmented inference was used where the loss function equals the sum of the $0$-$1$ losses per output variable, \ie, $L(\hat{y}, y)~\coloneqq~\sum_{i=1}^n~\mathbf{1}[\hat{y}_i~\neq~y_i]$ where $n$ is the number of output variables.

\paragraph{OCR:}
The Unary model is a single $3$-layer multilayer perceptron (MLP) with ReLU activations, hidden layer sizes of $200$, and a dropout layer after the first linear layer with keep probability $0.5$. Scores for each image were generated by independently passing them into this network. Both Struct and GSPEN use a graph with one pairwise region per pair of adjacent letters, for a total of $4$ pairs. Linear potentials are used, containing one entry per pair per set of assignments of values to each pair. The score function for both SPEN and GSPEN takes the form $F(x, p; w) = \sum_{r \in \mathcal{R}} \sum_{y_r \in \mathcal{Y}_r}p_r(y_r)b_r(x, y_;w) + T(B(x), p)$, where in the SPEN case $\mathcal{R}$ contains only unary regions and in the GSPEN case $\mathcal{R}$ consists of the graph used by Struct. Each $b_r$ represents the outputs of the same model as Unary/Struct for SPEN/GSPEN, respectively, and $B(x)$ represents the vector $(b_r(x, y_i;w)|_{y_r \in \mathcal{Y}_r})$. For every SPEN and GSPEN model trained, $T$ is a $2$-layer MLP with softplus activations, an output size of $1$, and either $500$, $1000$, or $2000$ hidden units. These hidden sizes as well as the number of epochs of training for each model were determined based on task performance on the validation data. Message-passing inference used in both Struct and GSPEN  ran for $10$ iterations. GSPEN models were trained by using the pre-trained Struct model, fixing these parameters, and training the $T$ function. The NLStruct model consisted of a $2$-layer MLP with $2834$ hidden units, an output size of $1$, and softplus activations. We use the same initialization described by \citet{graber2018nlstruct} for their word recognition experiments, where the first linear layer was initialized to the identity matrix and the second linear layer was initialized to a vector of all 1s. NLStruct models were initialized from the Struct models trained without entropy and used fixed potentials. The inference configuration described by \citet{graber2018nlstruct} was used, where inference was run for $100$ iterations with averaging applied over the final $50$ iterations.

All settings for the OCR experiments used a mini-batch size of $128$ and used the Adam optimizer, with Unary, SPEN, and GSPEN using a learning rate of $10^{-4}$ and Struct using a learning rate of $10^{-3}$. Gradients were clipped to a norm of $1$ before updates were applied. Inference in both SPEN and GSPEN were run for a maximum of $100$ iterations. Inference was terminated early for both models if the inference objective for all datapoints in the minibatch being processed changed by less than $0.0001$.

Three different versions of every model, initialized using different random seeds, were trained for these experiments. The plots represent the average of these trials, and the error represented is the standard deviation of these trials.

\paragraph{Tagging:}
The SPEN and GSPEN models for the image tagging experiment used the same scoring function form as for the OCR experiments. The $T$ model in both cases is a $2$-layer MLP with softplus activations and $130$ hidden units. Both GSPEN and SPEN use the same score function as in the OCR experiments, with the exception that the $T$ function used for GSPEN is only a function of the beliefs and does not include the potentials as input. Both models were trained using gradient descent with a learning rate of $10^{-2}$, a momentum of $0.9$, and a mini-batch size of $128$. Once again, only the $T$ component was trained for GSPEN, and the pairwise potentials were initialized to a Struct model trained using the settings described in \citet{graber2018nlstruct}.

The message-passing procedure used to solve the inner optimization problem for GSPEN was run for $100$ iterations per iteration of Frank-Wolfe.  Inference for SPEN and GSPEN was run for $100$ iterations and was terminated early if the inference objective for all datapoints in the minibatch being processed changed by less than $0.0001$.

\paragraph{Multilabel Classification:}
For the Bibtex dataset, $25$ percent of the training data was set aside to be used as validation data; this was not necessary for Bookmarks, which has a pre-specified validation dataset. For prediction in both datasets and for all models, a threshold determining the boundary between positive/negative label predictions was tuned on the validation dataset.

For the Bibtex dataset, the Unary model consists of a $3$-layer MLP taking the binary feature vectors as input and returning a $159$-dimensional vector representing the potentials for label assignments $y_i = 1$; the potentials for $y=0$ are fixed to $0$. The Unary model uses ReLU activations, hidden unit sizes of $150$, and dropout layers before the first and second linear layers with keep probability of $0.5$. The Struct model consists of a $2$-layer MLP which also uses the feature vector as input, and it contains $1000$ hidden units and ReLU activations. The SPEN model uses the same scoring function form as used in the previous experiments, except the $T$ function is only a function of the prediction vector and does not use the unary potentials as input. The $T$ model consists of a $2$-layer MLP which takes the vector $\left(p_i(y_i=1)\right)_{i=1}^{59}$ as input. This model has $16$ hidden units, an output size of $1$, and uses softplus activations. The GSPEN model was trained by starting from the SPEN model, fixing these parameters, and training a pairwise model with the same architecture as the Struct model. 

For the bookmarks dataset, the models  use the same architectures with slightly different configurations. the Unary model consists of a similar $3$-layer MLP, except dropout is only applied before the second linear layer. The Struct model uses the same architecture as the one trained on the Bibtex data. The $T$ model for SPEN/GSPEN uses $15$ hidden units.

For both datasets and for both SPEN and GSPEN, mirror descent was used for inference with an additional entropy term with a coefficient of $0.1$; for Struct, a coefficient of $1$ was used. Inference was run for $100$ iterations, with early termination as described previously using the same threshold. For Struct and GSPEN, message passing inference was run for $5$ iterations. The Unary model was trained using gradient descent with a learning rate of $10^{-2}$ and a momentum of $0.9$, while Struct, SPEN and GSPEN were trained using the Adam optimizer with a learning rate of $10^{-4}$.

\paragraph{NER:}
Both structured model baselines were trained using code provided by the authors of the respective papers. In both cases, hyperparameter choices for these structured models were chosen to be identical to the the choices made from their original works. For completeness, we will review these choices.

The structured model of \citet{ma2016end} first produces a vector for each word in the input sentence by concatenating  two  vectors: the first is a $100$-dimensional embedding for the word, which is initialized from pre-trained GloVe embeddings \citep{pennington2014glove} and fine-tuned. The second is the output of a $1$-D convolutional deep net with $30$ filters of length $3$ taking as input $30$-dimensional character embeddings for each character in the word. These representations are then passed into a $1$-layer bidirectional LSTM with a hidden state size of $256$, which is passed through a linear layer followed by an ELU activation to produce an intermediate representation. Unary/pairwise graphical model scores are finally obtained by passing this intermediate representation through two further linear layers. Predictions are made using the Viterbi algorithm. Dropout is applied to the embeddings before they are fed into the RNN (zeroing probability of $0.5511$, corresponding to two separate dropout layers with zeroing probability of $0.33$ being applied) and to the output hidden states of the RNN (zeroing probability of $0.5$). The GSPEN models in this setting were trained by initializing the structured component from the pre-trained models and fixing them -- that is, only the parameters in the MLP were trained during this step. Due to the fact that the input sentences are of varying size, we zero-pad all inputs of the MLP to the maximum sequence length. Dropout with a zeroing probability of $0.75$ was additionally applied to the inputs of the MLP. Inference was conducted using mirror descent with added entropy and convergence threshold of $0.1$. For both the structured baseline and GSPEN, model parameters were trained for $200$ epochs using SGD with initial learning rate of $0.01$, which was decayed every epoch using the formula $\text{lr}(\text{epoch})~=~\tfrac{0.01}{1+\text{epoch}\cdot0.05}$. The structured baseline was trained with a mini-batch size of $16$, while the GSPEN model used a mini-batch size of $32$ during training. A larger batch size was used for the GSPEN model to decrease the amount of time to complete one pass through the data.

\citet{akbik2018contextual} use a concatenation of three different pre-trained embeddings per token as input to the bidirectional LSTM. The first is generated by a bidirectional LSTM which takes character-level embeddings as input and is pre-trained using a character-based language modeling objective (see \citet{akbik2018contextual} for more details). The other two embeddings are GloVe word embeddings \citep{pennington2014glove}, and task-trained character-based embeddings (as specified by \citet{lample2016neural}). During training, these embeddings are fine-tuned by passing them through a linear layer whose parameters are learned. The embeddings are passed into a $1$-layer bidirectional LSTM with a hidden state size of $256$. Unary scores are generated from the outputs of the LSTM by passing them through a linear layer; pairwise scores consist of a matrix of scores for every pair of labels, which are shared across sentence indices. In this setting, the GSPEN models were trained by initializing the structured component from the pre-trained models and then fine-tuning all of the model parameters. Mirror descent with added entropy was used for GSPEN inference with a convergence threshold of $0.1$. For both the structured baseline and GSPEN, model parameters were trained for a maximum of $150$ epochs using SGD with mini-batch size of $32$ and initial learning rate of $0.1$, which was decayed by $0.5$ when the training loss did not decrease past its current minimum for $3$ epochs. Training was terminated early if the learning rate fell below $10^{-4}$.

\end{document}